\documentclass[conference]{IEEEtran}  

\IEEEoverridecommandlockouts                              

\usepackage{amsmath} 
\usepackage{amssymb}  
\usepackage{todonotes}
\usepackage{enumitem}
\usepackage{mathtools}
\usepackage{rotating}
\usepackage{multirow}
\usepackage{siunitx}
\usepackage{caption}
\usepackage{subcaption}
\usepackage{float}
\usepackage{hyperref}
\usepackage{pifont}
\usepackage{cite}

\usepackage{booktabs} 

\usepackage{algorithmic} 
\usepackage{algorithm2e}
\graphicspath{{graphics/}}

\usepackage{tabto} 

\usetikzlibrary{matrix, positioning, calc}
\tikzset{
  quadratic/.style={
    to path={
      (\tikztostart) .. controls
      ($#1!1/3!(\tikztostart)$) and ($#1!1/3!(\tikztotarget)$)
      .. (\tikztotarget)
    }
  }
}

\hypersetup{
    colorlinks=true,
    linkcolor=black,
    filecolor=magenta,      
    urlcolor=cyan,
    pdftitle={Overleaf Example},
    pdfpagemode=FullScreen,
    }

\title{\LARGE \bf
EPSM: A Novel Metric to Evaluate the Safety of Environmental Perception in Autonomous Driving 
}

\author{Jörg Gamerdinger, Sven Teufel, Stephan Amann, Lukas Marc Listl, and Oliver Bringmann
\thanks{University of T\"ubingen, Faculty of Science, Department of Computer Science, Embedded Systems Group 
\tt\small {\{joerg.gamerdinger, sven.teufel, stephan.amann, lukas.listl, oliver.bringmann\} @uni-tuebingen.de}}
}%

\begin{document}
\maketitle
\thispagestyle{empty}
\pagestyle{empty}

\begin{abstract}

Extensive evaluation of perception systems is crucial for ensuring the safety of intelligent vehicles in complex driving scenarios. Conventional performance metrics such as precision, recall and the F1-score assess the overall detection accuracy, but they do not consider the safety-relevant aspects of perception. Consequently, perception systems that achieve high scores in these metrics may still cause misdetections that could lead to severe accidents.
Therefore, it is important to evaluate not only the overall performance of perception systems, but also their safety. We therefore introduce a novel safety metric for jointly evaluating the most critical perception tasks, object and lane detection. Our proposed framework integrates a new, lightweight object safety metric that quantifies the potential risk associated with object detection errors, as well as an lane safety metric including the interdependence between both tasks that can occur in safety evaluation. The resulting combined safety score provides a unified, interpretable measure of perception safety performance. Using the DeepAccident dataset, we demonstrate that our approach identifies safety critical perception errors that conventional performance metrics fail to capture. Our findings emphasize the importance of safety-centric evaluation methods for perception systems in autonomous driving.

\end{abstract}

\section{INTRODUCTION}
\label{sec:intro}
 
The most likely cause of death for young people aged 5 to 29 is road traffic injury~\cite{Death}. Particularly in adverse weather conditions such as rain and snow, the rate of accidents and injuries can increase by \SIrange{70}{80}{\percent}. Automated vehicles are a promising approach to reducing accident rates, as human error is the main cause fatal road accidents~\cite{EuropeanUnion2019}. 
In order to achieve safe automated driving, a comprehensive perception of the environment is crucial for automated vehicles. This includes not only object detection but also lane detection, which is necessary for safe trajectory and motion planning. However, in benchmarks, these detection methods are evaluated using state-of-the-art performance metrics such as accuracy, recall, F1-score or average precision that do not reflect the safety of the perception~\cite{volk2020safety}. Hence, appropriate metrics are required for the offline evaluation of perception algorithms. For offline evaluation, as used in benchmarks, ground truth information such as a map information and vehicle states are known. This is in contrast to an online evaluation, in which the safety is not evaluated but estimated while driving. 

State-of-the-art safety evaluation approaches only consider the independent evaluation of perception tasks. 
However, tasks solved in the perception of intelligent vehicles can have a strong interdependence when it comes to the safety in a scenario. The two most safety critical tasks in perception, namely lane detection and object detection influence each other in terms of safety. For example, the misclassifications of a lane can influence the criticality of an object in a nearby lane.
Therefore, we present our environmental perception safety metric EPSM that jointly evaluates lane detection and object detection to cover the most critical tasks in autonomous driving perception.\\

\begin{figure}[t]
    \centering
    \includegraphics[width=.9\linewidth, page=1, trim= 18cm 3.5cm 18cm 3.2cm, clip]{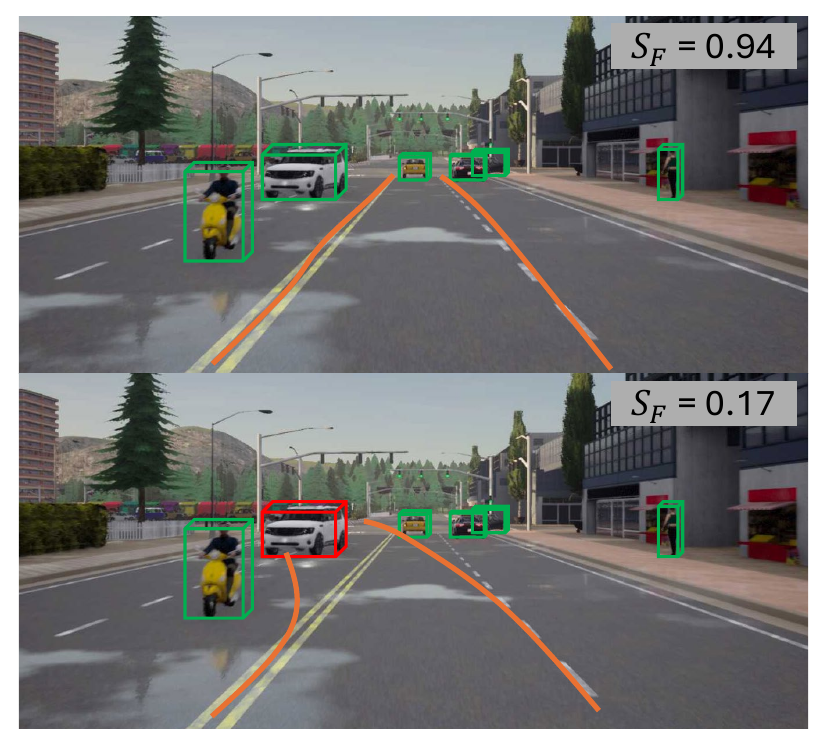}
    \caption{Exemplary scenes including the resulting safety score $S$. Green boxes mark correct detections, red represents an object which is not detected. Orange lines represent the detected lane.}
    \label{fig:example}
    \vspace*{-6mm}
\end{figure}
\vspace*{-2mm}
Our main contributions are:
\begin{itemize}[noitemsep]
    \item We propose a novel light-weight safety metric, including an optimized criticality classification, to evaluate safety in object detection\\[-2mm]
    \item We present the first single metric for the joint safety assessment of lane and object detection systems, including the interdependence between both tasks in safety evaluation\\[-2mm]
    \item We demonstrate the advantages of our metric compared to common performance metrics in safety-critical situations
\end{itemize}
\ \\
In Sec.~\ref{sec:related_work} we provide an overview of existing methods for performance and safety evaluation for environmental perception. Section~\ref{sec:method} describes our joint metric for a the safety evaluation of object and lane detection. The results are presented in Sec.~\ref{sec:results}, followed by a conclusion and an outlook.

\section{RELATED WORK}
\label{sec:related_work}

There exist multiple performance metrics to evaluate object and lane detection algorithms. 

Most of these metrics represent the ratios of correctly detected objects or lane points. For this, True positive (TP) describes a correctly detected pixel or a correctly detected object. False positive (FP) stands for a pixel which is erroneously classified as lane pixel or an object that does not exist. False negative (FN) describes a pixel of the lane which is not classified as lane or an object that is not detected.
To determine if a lane point is classified as TP / FP, 2D or 3D distances with a threshold are used. If the detection is closer to the ground truth (GT) as the threshold, it is counted as TP.

For object detection mostly the Intersection over Union (IoU) also known as Jaccard-Index is used to classify between TP and FP. It uses the area (2D) or volume (3D) of the intersection and the union of a detected bounding box $D$ and the corresponding GT bounding box $G$ and is defined as~\cite{IoU}

\begin{equation}
\label{eq:iou}
\mathrm{IoU} = \frac{|D \cap G|}{|D \cup G|}.
\end{equation}

The commonly used thresholds for IoU to classify detections as TP are 0.5 or 0.7~\cite{kitti}.

Using these classifications, various performance metrics can be calculated.
The precision $P$ and recall $R$ for evaluation as defined in Eq.~\eqref{eq:prec}: 
\begin{equation}
    \label{eq:prec}
    P = \frac{TP}{TP+FP},\qquad R = \frac{TP}{TP+FN},
\end{equation}
are the most simple metrics used by state-of-the-art lane detection benchmarks~\cite{TuSimple, CULane, Fritsch2013ITSC, LLamas} on a 2D pixel level.
A further metric is the accuracy $A$ as defined in Eq.~\eqref{eq:accuracy} which is used in the TuSimple benchmark~\cite{TuSimple}.
\begin{equation}
\label{eq:accuracy}
\text{A} = \frac{TP + TN}{TP + TN + FP + FN}.
\end{equation}

As a harmonic mean of precision and recall the $F1$-score can be derived. $F1$ is designed to incorporate the advantages of both metrics.
\begin{equation}
    \label{eq:f1}
    F1 = 2\cdot\frac{P\cdot R}{P+R}
\end{equation}

In object detection, mostly average precision (AP) or mean average precision (mAP) as defined in Eq.\eqref{eq:ap} and Eq.\eqref{eq:map} are used~\cite{everingham2010pascal}.
\begin{equation}
\label{eq:ap}
    AP_c = \int_{0}^{1} P_c(R) \, dR
\end{equation}
\begin{equation}
\label{eq:map}
    mAP = \frac{1}{N} \sum_{c=1}^{N} \text{AP}_c
\end{equation}
Here, $R$ represents the recall and $P(R)$ is the precision as a function of recall. While the $AP$ is calculated for a specific object type $c$, the mAP is the mean of $AP$ for all classes.

For an easier calculation, different benchmarks use interpolated versions such as the 11-point interpolated AP used in Pascal VOC~\cite{everingham2010pascal} and KITTI~\cite{kitti}, as shown in Eq.~\eqref{eq:11pap}.
\begin{equation}
    \label{eq:11pap}
AP_c^{11} = \frac{1}{11} \sum_{r \in \{0.0, 0.1, \dots, 1.0\}} 
\max_{r' \ge r} p_c(r')
\end{equation}

A more accurate version called all point interpolated AP is used by Waymo~\cite{waymo} and is defined as shown in Eq.~\eqref{eq:apap}.

\begin{equation}
\begin{aligned}
AP_c &= \sum_{i=1}^{m-1} (r_{i+1} - r_i) \, p_{\mathrm{interp}}(r_{i+1}), \\
p_{\mathrm{interp}}(r) &= \max_{r' \ge r} p(r')
\end{aligned}
\label{eq:apap}
\end{equation}

All these metrics have the disadvantage that they neglect semantic information such as object types or velocities. Hence, in research some improved metrics were presented.

For lane detection, Fritsch et al.~\cite{Fritsch2013ITSC} proposed a behavior-based method where the driving corridor with the highest probability based on the detected boundaries is evaluated using precision and recall. 
Gamerdinger et al.~\cite{gamerdinger2023cold} as well as Sato et al.~\cite{sato2022towards} used the mean and maximum deviation in meter as evaluation metric, which leads to a more meaningful result in many scenarios, but still cannot assess the safety as information about the vehicle and the scene are not taken into account. 

For object detection, Stiefelhagen et al.~\cite{Stiefelhagen_et_al_CLEAR} proposed the metrics Multiple-Object-Detection Precision (MODP) and Accuracy (MODA), as defined in Eq.~\eqref{eq:MODA} and Eq.~\eqref{eq:MODP}.
$m_t$ and $\mathit{fp}_t$ be the amount of misses and false positives at time $t$ and $g_t$ be the number of ground truth objects at time $t$. The MODP score is determined by all mapped object sets which are used to calculate the IoU of each object. $N^{\mathrm{mapped}}_t$ represents the number of mapped object sets at $t$.

\begin{equation}
\label{eq:MODA}
\mathrm{MODA}(t)=1-\frac{\sum_t (m_t + \mathit{fp}_t)}{\sum_t g_t},
\end{equation}

\begin{equation}
\label{eq:MODP}
\mathrm{MODP}(t)=\frac{\sum_{i=1}^{N^{\mathrm{mapped}}_t} \mathrm{IoU}_{i}}{N^{\mathrm{mapped}}_t}.
\end{equation}

Using the IoU to determine the precision score allows a better statement about the precision compared to the binary classification and calculation based on TP and FP amount. However, it still neglects safety relevant information such as velocities or object types.

Kim et al.~\cite{RTP16} proposed a metric for real-time video surveillance systems; they include the time which is required to detect an object. In the proposed soft-real-time mode the CLEAR metric scores are mapped to an interval of $[0;\mathrm{score}]$ if detection time exceeds a given threshold $\tau$ while in a hard-real-time mode in which the score is set to 0 if detection time exceeds $\tau$. 
However, in the field of video surveillance the relevance factor is less essential than in automated driving. The extension of the CLEAR metrics of Kim et al. cannot, therefore, be directly applied to evaluate perception algorithms for autonomous vehicles. 

Hence, a first comprehensive safety metric (CSM) for object detection in automated driving was presented by Volk et al.~\cite{volk2020safety}. This metric allows for a comprehensive safety assessment of object detection algorithms by incorporating safety relevant information and scene semantics such as velocity and the object type. As a basis they used the CLEAR metrics which are then adapted by a time factor as proposed by Kim et al.~\cite{RTP16} and a collision factor determined by the severity of a potential collision if the object is not correctly detected. This metric can be applied to object detection and tracking, but not for lane detection. Additionally, some of their intermediate metrics correlate and the employed RSS metric to determine the perception relevance does not provide optimal results in safety-critical situations~\cite{gamerdinger2025criticality}.

\begin{figure*}[t]
    \centering
    \includegraphics[width=\linewidth, page=6, trim= 0cm 7cm 0cm 0cm, clip]{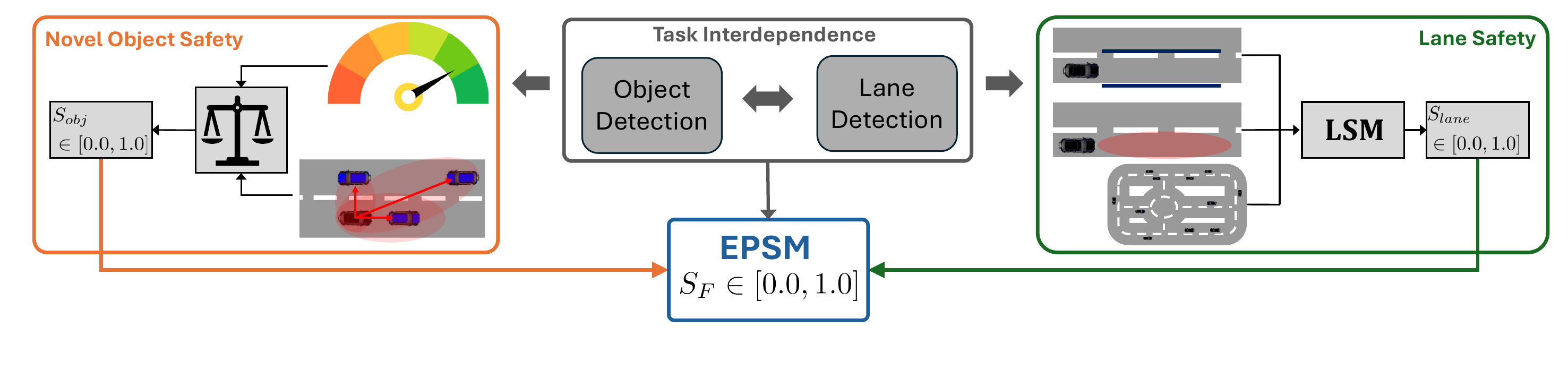}
    \caption{Overview of the proposed EPSM and the intermediate metrics.}
    \vspace*{-4mm}
    \label{fig:overview}
\end{figure*}

Considering lane detection, a first metric for safety evaluation was proposed by Gamerdinger et al.~\cite{gamerdinger2024lsm}. This lane safety metric (LSM) incorporates safety relevant factors such as the lane type and velocities to determine the safety of the lane detection. The metric uses a required longitudinal detection distance, the lateral detection deviation and scenario semantics such as the type of adjacent lanes and speed limits to provide a single safety score that can be divided into five categories for a better interpretability of the result. However, the metric only considers the safety of the lane detection and does not incorporate any further aspects about other traffic participants which can be relevant in terms of safety.

\section{PERCEPTION SAFETY METRIC}
\label{sec:method}
\subsection{Lightweight Object Safety Metric}
\label{subsec:object_metric}

Factors like detection time and collision occurrence as used in the CSM by Volk et al.~\cite{volk2020safety}, heavily correlate and therefore must not be considered both. In addition, the tracking result strongly correlates with the detection results as the detection data serves as input for the tracking. Hence, our metric removes the tracking component and focuses on the detection which is more crucial. In order to achieve a reliable and comprehensible safety assessment we include a novel lightweight object safety metric which consists of three main components: Criticality, Severity and a weighting based on criticality and severity. These main components are explained in the following.

\paragraph{Criticality}
The criticality of an object, describes its necessity to perceive it to avoid a safety-critical situation~\cite{gamerdinger2025criticality}. As an example, a pedestrian crossing the road in front of the ego vehicle must be perceived, while a vehicle driving \SI{50}{\meter} behind the ego in opposite direction does not affect safety if it is not perceived. Different metrics exist that allow for such a classification of criticality. However, as shown in~\cite{gamerdinger2025criticality} single metrics do not generalize well across different scenarios. Hence, we use multi-metric aggregation to overcome limitations of single metrics.
We consider three metrics to cover both collision and near collision scenarios as well as vehicles behind the ego.
The first used metric is the Time-to-Collision (TTC) as introduced by Hayward~\cite{hayward1972near}. This metric determines the time until the trajectories of two objects intersect. In literature, different threshold values between \SI{1}{\second} and \SI{4}{\second} are used as thresholds to classify between critical or not. However, we aim for a continuous criticality score $C\in [0.0,1.0]$ for a more meaningful statement about the criticality. Hence, we map the TTC to the TTC criticality score $C_{TTC}$ using Eq.~\eqref{eq:ttc_mapping}.

\begin{equation}
\label{eq:ttc_mapping}
    C_{TTC} = \frac{1}{1+e^{k\cdot(TTC - t_{falloff})}}
\end{equation}
With $k=3$ and $t_{falloff} = 2.5\si{\second}$ this leads to $C_{TTC} = 1.0$ for $TTC = \SI{1}{\second}$ and $C_{TTC} = 0.0$ for $TTC = \SI{4}{\second}$ which corresponds to the thresholds used in literature~\cite{gamerdinger2025criticality}.

As the TTC only classifies colliding objects as critical, we further incorporate the Time-To-Closest-Encounter (TTCE)~\cite{hayward1972near}. This metric is particularly relevant for near-miss situations, where no collision occurs but the objects pass in close proximity. The mapping from TTCE to the criticality score $C_{TTCE}$ is performed using Eq.~\eqref{eq:ttc_mapping}. In addition, the TTCE also calculates the minimum distance between the objects $d_{TTCE}$. Using $d_{TTCE}$ we create a mapping similar to the time mapping in Eq.~\eqref{eq:ttc_mapping}. The mapping is performed using Eq.~\eqref{eq:ttce_mapping}

\begin{equation} \label{eq:ttce_mapping}
\begin{split}
 &C_{TTCE} = max\\
 &(\frac{1}{1+e^{k\cdot(TTCE - t_{falloff})}}, \\
    &\frac{1}{1+e^{k\cdot(d_{TTCE} - d_{falloff})}}),
\end{split}
\end{equation}
    with $k=3$ and $d_{falloff} = 4.5\si{\metre}$ which roughly corresponds to the length of a regular car.

From the ego's point of view, the TTC and TTCE in various cases only consider safety-critical situations in driving direction (mostly front) of the ego vehicle. Hence, as a third metric for rear end collisions is required. To incorporate these scenarios the bidirectional rating as proposed in~\cite{gamerdinger2025criticality} is used and applied to the vehicle located behind the ego vehicle by using Eq.~\eqref{eq:ttc_mapping} and Eq.~\eqref{eq:ttce_mapping} to obtain the rear end criticality score $C_{REAR}$.

The resulting final criticality score $C_o$ for an non-VRU object is determined using Eq.~\eqref{eq:c_o}.
\begin{equation}
\label{eq:c_o}
    C_o = max (C_{TTC}, C_{TTCE}, C_{REAR})
\end{equation}

Vulnerable road users (VRUs), especially pedestrians, have less constraints in their movement compared to vehicles as they can abruptly change their direction of movement by any degree. To account for this, the pedestrian criticality is modeled using a circular criticality zone with a diameter of eight times the speed of the VRU. This matches the upper threshold of $\SI{4}{\second}$ for TTC as used in literature. Due to the lower predictability of the VRU movement and the increased severity, the criticality for a VRU $C_{VRU}$ is set to 1.0 if the distance to the VRU is lower than the criticality zone radius which is about $\SIrange{5}{6}{\meter}$ for a normal walking speed of pedestrians.

\paragraph{Severity}
Besides the criticality, the severity of a potential collision in case of a missing detection is the second key aspect in safety evaluation. The severity depends on multiple factors such as velocity, age and object type. Since VRUs like pedestrians or cyclists have no deformation zone, they show a higher vulnerability compared to passengers in a vehicle. Hence, the severity score $I$ which represents the impact of the collision is determined with respect to the object type.

For VRUs the logistic regression model by Saadé et al.~\cite{saade2020pedestrian} is employed. The model was created using a large-scale real-world pedestrian accident dataset. It incorporates the velocity $V$ of the vehicle and the age $A$ of the pedestrian to determine the probability of a fatality $P_K$ using Eq.~\eqref{eq:p_k} and the probability of a serious injury using Eq.~\eqref{eq:p_ksi}.

\begin{equation}
\label{eq:p_k}
P_K = \frac{\exp(-8.0941 + 0.0012 \cdot V^2 + 0.0525 \cdot A)}{1 + \exp(-8.0941 + 0.0012 \cdot V^2 + 0.0525 \cdot A)}
\end{equation}

\begin{equation}
\label{eq:p_ksi}
P_{KSI} = \frac{\exp(-2.9893 + 0.0013 \cdot V^2 + 0.0286 \cdot A)}{1 + \exp(-2.9893 + 0.0013 \cdot V^2 + 0.0286 \cdot A)}
\end{equation}
To obtain a usable severity score, it is necessary to map probabilities onto severity. Hence, we create a mapping for the probabilities from Eq.~\eqref{eq:p_k} and Eq.~\eqref{eq:p_ksi} to an interval $[0.0, 1.0]$ with respect to the five safety levels used in our metric (see Tab.~\ref{tab:safety_classification}). If $P_K\geq0.5$, we consider a fatal collision as most likely which corresponds to the highest severity. The mapping is performed using a linear mapping as shown in Eq.~\eqref{eq:lin_mapping} with $p = P_K$. Moreover, $p_{min}=0.5, p_{max}=1.0, I_{min}=0.8, I_{max}= 1.0$ hold as the probability between 0.5 and 1.0 is mapped to $I \in [0.8, 1.0]$.
Equivalent for $P_{KSI} \geq 0.5$, a mapping to the interval $[0.4, 0.8)$ is performed using Eq.~\eqref{eq:lin_mapping} with $p=P_{KSI}$. For $P_{KSI} < 0.5$, no harming effects on the VRU are predicted. This means we are either in a safe status, in which no severity is present or only minor bodywork damage can occur. Hence, the probability is mapped to $[0.0, 0.4)$.

\begin{equation}
    \label{eq:lin_mapping}
    I = I_{min} + \frac{(p - p_{min})(I_{max} - I_{min})}{p_{max}-p_{min}}
\end{equation}

Considering vehicles, the model by Malliaris et al.~\cite{malliaris1997relationships} is employed. This model was fitted on a large-scale accident dataset from the United States. The model incorporates relative velocity and impact direction in order to distinguish between different types of accidents. The logisitic regression model determines the probability $P$ of a collision being classified as \textit{MAIS 2+}, \textit{MAIS 3+} or \textit{fatal}. The \textit{MAIS} classification was introduced by Rapsang and Shyam~\cite{rapsang2015scoring}. \textit{MAIS 2+} corresponds to moderate injuries such as simple fractures while \textit{MAIS 3+} represents serious injuries as reported in safety and public health statistics.
If $0.5 \leq P_{fatal} \leq 1.0$, a linear mapping of the severity score $I$ to $[0.8,1.0]$ is applied. This corresponds to the inverse of the safety score rating for fatality. For $0.5 \leq P_{\textit{MAIS 3+}} \leq 1.0$ the mapping is applied within an interval of $I\in [0.2, 0.8)$ and for $0.5 \leq P_{\textit{MAIS 2+}}\leq 1.0$ which corresponds to a minor severity $I\in [0.0,0.2]$ holds.

\paragraph{Weighting}
Similar to the approach used to determine ASIL levels~\cite{ISO26262-3-2018} for automated driving, we consider severity and probability, which, in this case, is represented by the criticality. For each object $o$ we determine a weight $W_o$ using Eq.~\eqref{eq:weighting} with the corresponding criticality $C_o$ and the severity $I_o$.
\begin{equation}
    \label{eq:weighting}
    W_o = C_o \cdot I_o
\end{equation}

To account for the fact that some safety-critical misdetections have a stronger influence to safety than multiple good detections, the worst case is weighted higher. For this purpose we sort the weights together with the corresponding criticalities in increasing order. Using Eq.~\eqref{eq:object_safety_score} the weighted safety score for the object detection $S_{obj}$ is determined. 

\begin{equation}
\label{eq:object_safety_score}
S_{obj} = 1 - \frac{16 \cdot w_{0} + 4 \cdot w_{1} + \sum_{i=2}^{n} w_{i}}
{16 \cdot c_{0} + 4 \cdot c_{1} + \sum_{i=2}^{n} c_{i}}
\end{equation}
where $n$ represents the number of objects in the environment.

\subsection{Lane Safety Metric}
\label{subsec:lane_metric}

The lane perception safety metric used in this work is based on the LSM by Gamerdinger et al.~\cite{gamerdinger2024lsm} and consists of three components: (1) a longitudinal safety rating assessing the detection range, (2) a lateral safety rating measuring deviation between detected and ground-truth lanes, and (3) a scenario semantic rating evaluating the criticality of deviations depending on the driving context.

Lane boundaries are represented as linestrings defining the detected lane. Additional inputs include braking acceleration, vehicle and lane dimensions, and map information. For this work, the lateral safety rating was adapted by computing the mean lateral deviation
\[
d_{\text{lat,mean}} = \frac{1}{N}\sum_{i=1}^{N} d_{\text{lat}}(i)
\]
between corresponding points on the detected and ground-truth centerlines. This accounts for both near-field and far-field deviations, providing a balanced measure of lateral accuracy. The mean deviation is compared to a threshold $th_{\text{lat}}$ derived from the lane-to-vehicle width ratio. If $d_{\text{lat,mean}} < 0.8\cdot th_{\text{lat}}$, the lateral score is set to $S_{\text{lat}} \in (0.8, 1.0]$; otherwise, $S_{\text{lat}} = 0.8$, which triggers the scenario semantic rating.

The scenario semantic rating incorporates the impact velocity of a potential opposing-lane collision when the detected lane intrudes into an adjacent oncoming lane. The resulting score $S_{\text{sem}} \in [0.0, 0.8]$ is determined from the relative velocity between the ego vehicle and oncoming traffic and categorized into four severity levels~\cite{gamerdinger2024lsm}.

Finally, the three ratings are combined into a single lane safety score
\[
S_{\text{lane}} \in [0.0, 1.0],
\]
following the aggregation method as proposed in~\cite{gamerdinger2024lsm}.

\subsection{Combined Safety Score and Task Interdependence}
As the goal of this work is to combine the safety evaluation of object and lane detection within one single and easily comparable metric, the result of the evaluation must be a single score. Similar to commonly used performance metrics or existing safety metrics~\cite{volk2020safety, gamerdinger2024lsm}, our metric outputs a single final score $S_F\in[0.0,1.0]$. 

\begin{figure}[h!]
    \centering
    \includegraphics[width=.9\linewidth]{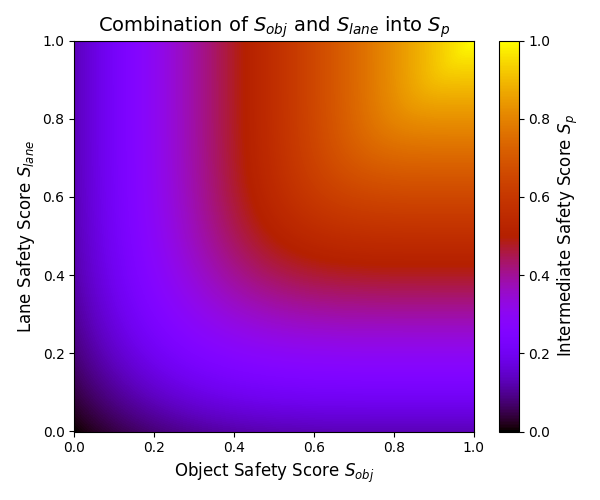}
    \caption{Visualization of power mean function to combine intermediate metric scores $S_{obj}$ and $S_{lane}$ into the single intermediate score $S_p$.}
    \label{fig:power_mean}
\end{figure}

To combine the two intermediate metric safety scores from the object detection $S_{obj}$ (see  Sec.~\ref{subsec:object_metric}) and the lane detection $S_{lane}$ (see Sec.~\ref{subsec:lane_metric}) into a single intermediate score $S_p$, we employ a power mean function as defined in Eq.~\eqref{eq:power_mean}. The influence of different inputs within the power mean function is shown in Fig.~\ref{fig:power_mean}.

\begin{equation}
\label{eq:power_mean}
S_p = 1 - \left( \frac{(1 - S_{obj})^p + (1 - S_{lane})^p}{2} \right)^{\frac{1}{p}}
\end{equation}

\begin{figure}
    \centering
    \includegraphics[width=.9\linewidth, page=7, trim= 15cm 4cm 15cm 4cm, clip]{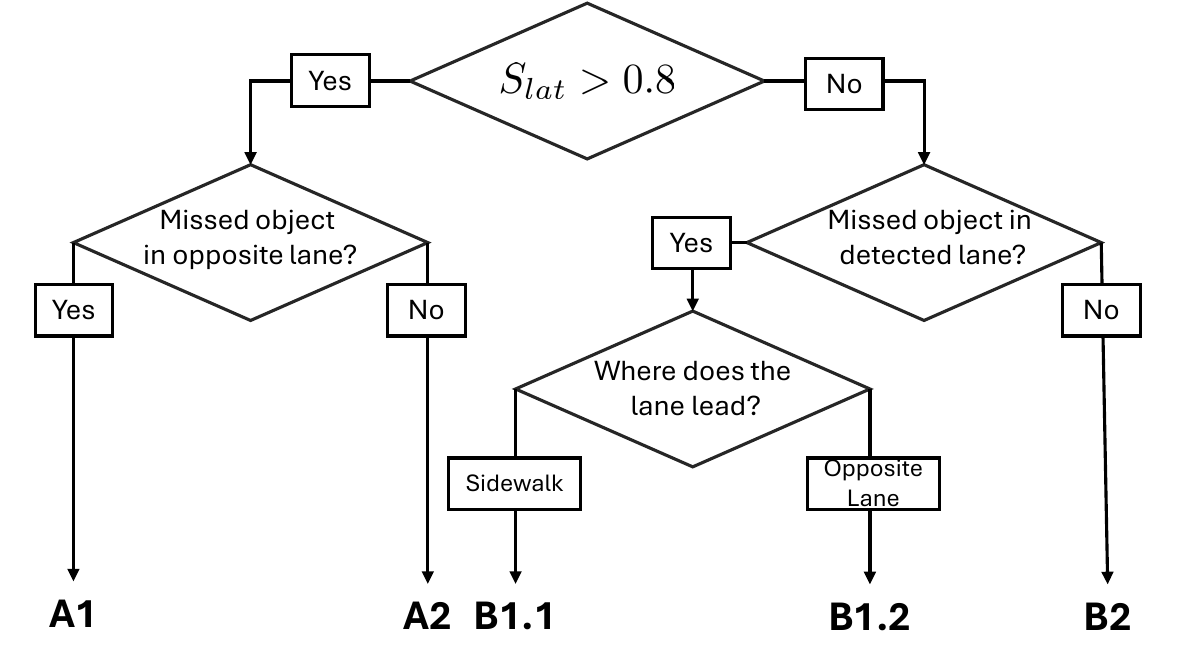}
    \caption{Visualization of the decision tree to fine-tune the intermediate safety score $S_p$}
    \label{fig:decision_tree}
    \vspace*{-5mm}
\end{figure}

This intermediate safety score $S_p$ is fine-tuned in a last step based on a decision tree (see Fig.~\ref{fig:decision_tree}) which represents different situations in which an interdependence between the object and lane detection evaluation occurs.

Based on the lateral safety score of the LSM, which describes if the vehicle is able to stay on it's own lane, a distinction between cases A and B is done. 
The case A represents, a safe lateral detection of the lane. In subcase A1, in the adjacent lane is a missed object; however, as the lane is safely perceived there is no negative affection to the safety which leads to a minor bonus. If no object is present, no influence on the safety score is present and there is no action (A2). 
The second case (B) represents a unsafe lateral detection. In the second level the decision is taken based on if a missed object is within the detected lane. If no (B2), there is no negative influence on the safety and as in A1 a bonus is applied. 
In the case that there is an undetected object within the lane, a distinction between roads (B1.2) and sidewalks (B1.1) is taken. This is done to differentiate between possible collisions with other vehicles or VRUs which are moving on sidewalks. In both cases a penalty is applied to $S_p$.

The bonus and penalty for the different cases is determined as follows:
\begin{itemize}
    \item \textbf{A1 + B2:} As the appropriate detection of the lane can avoid a collision. This leads to a bonus factor $F_b$ which is combined with the intermediate safety score $S_p$ to determine the final safety score $S_F$ as shown in Eq.~\eqref{eq:bonus}. 
    \begin{equation}
    \label{eq:bonus}
S_F =
\begin{cases}
S_p + (F_b - 1)\cdot(1 - S_I), & \text{if } F_b > 1 \\
S_p - (1 - F_b)\cdot S_I, & \text{if } F_b < 1 \\
S_p, & \text{else}
\end{cases}
\end{equation}

For increasing $S_p$, the effect of the bonus is lower. Hence, the bonus does not indicate high safety if it is not present. Also, it does not affect an intermediate safety of $S_p = 0.0$.

    \item \textbf{B1.1 + B1.2:} In both cases, a penalty factor $F_p$ determined by distance and relative velocity is defined. For this purpose, the lowest Time-to-Collision $TTC_{min}$ of all missed objects is taken into account. Based on this, the final safety score $S_F$ is determined as follows:
    \begin{itemize}
    \item \( TTC_{\min} \leq 2s \rightarrow S_C \) is linearly mapped to \([0.6, 0.8]\)
    \item \( TTC_{\min} \leq 4s \rightarrow S_C \) is linearly mapped to \([0.8, 0.9]\)
    \item \( TTC_{\min} \leq 8s \rightarrow S_C \) is linearly mapped to \([0.9, 1.0]\)
\end{itemize}
\end{itemize}

In contrast to the performance metrics, the result of the safety metric is more difficult to interpret and compare. To overcome this drawback, we use the five level classification as shown in Tab.~\ref{tab:safety_classification}.

\begin{table}[h!]
\centering
\caption{Adapted safety metric score classification from~\cite{volk2020safety}.}
\begin{tabular}{c    l} \toprule
    {$\textit{\textbf{$S_F$}} \boldsymbol{\in}$} & {$\textit{\textbf{Classification}}$} \\ \midrule
    \multicolumn{1}{r}{{[}0.0 - 0.2{]}}  & \textbf{insufficient}, high risk of fatality \\
    \multicolumn{1}{r}{{(}0.2 - 0.4{]}} & \textbf{very bad}, existing risk for serious violation \\
    \multicolumn{1}{r}{{(}0.4 - 0.6{]}}  & \textbf{bad}, low probability of minor injuries  \\
    \multicolumn{1}{r}{{(}0.6 - 0.8{]}} & \textbf{good}, low risk of bodywork damage    \\
    \multicolumn{1}{r}{{(}0.8 - 1.0{]}} & \textbf{very good}, high probability of safe status  \\ \bottomrule
\end{tabular}
\vspace{-1mm}
\label{tab:safety_classification}
\end{table}

\section{EXPERIMENTS AND RESULTS}
\label{sec:results}
In order to provide a meaningful evaluation of the proposed safety metric, it is necessary to incorporate safety-critical scenarios. Wang et al.~\cite{wang2023deepaccident} proposed the DeepAccident dataset, a diverse dataset of 57k frames designed for accident prediction and collective perception. The dataset comprises approximately 690 scenarios, encompassing both accident and non-accident scenarios. 
The evaluation of criticality metrics is facilitated by all scenarios that incorporate a safety-critical situation, defined in this case by an accident. Given that the dataset only provides complete meta information within the train data, 159 scenarios with a total of 4995 frames were extracted from the train set, incorporating the IDs of colliding objects. 
In each scenario two vehicles, one denoted as ego and a second object collide. 

In order to provide meaningful results, which allow to demonstrate the advantage of our metric compared to commonly used performance metrics, we designed two controllable and statistical sensor models for object and lane detection. This allows to control the perception capabilities and to create misdetections leading to safety-critical scenarios demonstrating the behavior of the metric with varying inputs. It must be noted that this does not limit the applicability of the metric to state-of-the-art neural network based object and lane detectors.

For map and lane processing the Lanelet2 framework is used~\cite{lanelet2}. The framework provides multiple functions for handling and querying maps as well as additional modules for routing or the representation of traffic rules. The lane detection model uses the provided ground truth map to sample a lane as described in~\cite{gamerdinger2023cold}. We can set a maximum distance which is covered by the detection model, however; also occlusion by curves are included and limit the perception range. Moreover, a normal distributed error $\delta_{lane} \sim \mathcal{N}(0, 0.05)$ is applied to simulate real-world perception inaccuracies.
Similarly, we use a statistical \SI{360}{\degree} LiDAR sensor model incorporating the distance depending detection probabilities from~\cite{zhou2020end, hu2022point}. They investigated the detection performance depending on the distance from the sensor which allows for a sufficient approximation of a detection model. In addition, a randomly sampled error to the size of the bounding box or the orientation is applied, which increases the realism of the model. As for the lane detector this allows for a reproducible and controllable perception to demonstrate the behavior of the proposed metric.

The metric-specific parameters used for this evaluation, shown in Tab.~\ref{tab:experimental_parameters}, are based on large empirical ablations studies.

\begin{table}[t]
\centering
\caption{Parameters used for the Experimental Setup.}
\begin{tabular}{ll}
\hline
\textbf{Parameter} & \textbf{Value} \\
\hline
Power Mean p & 5\\
IoU threshold Vehicle & 0.7 \\
IoU threshold VRU & 0.5 \\
Detection Distance & 50\,m \\
Bonus factor $F_b$ & 1.1 \\
\hline
\end{tabular}
\label{tab:experimental_parameters}
\end{table}

\subsection{Limitations of Performance Metrics}
\begin{figure*}[t]
    \centering
    \includegraphics[width=\linewidth, page=4, trim= 8cm 4cm 8cm 4cm, clip]{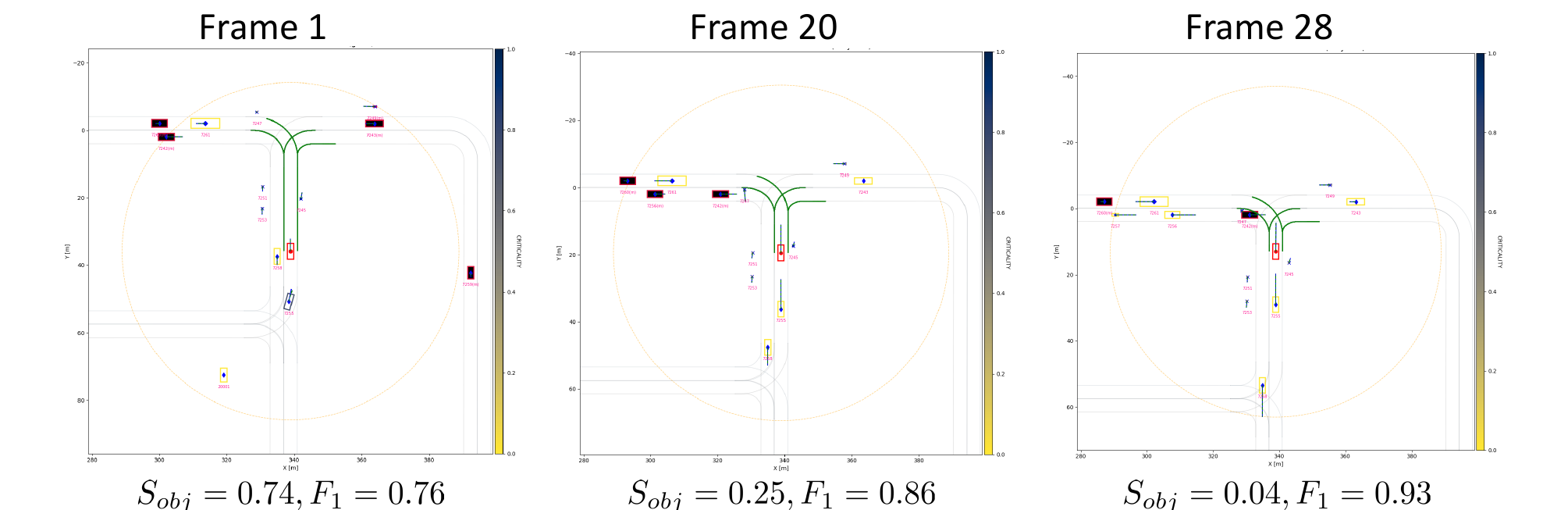}
    \caption{Example scenario, showing the advantage of the proposed object safety metric compared to common performance metrics. Red vehicle in the center represents the ego with its detected lane (green). Other red boxes with filling represent undetected objects with their criticality.}
    \vspace*{-0mm}
    \label{fig:safety_scenario}
\end{figure*}

Within this section, an exemplary scenario is shown to demonstrate the function of the novel lightweight object safety metric. While for most scenarios a strong correlation between the performance and safety exists, there exist various scenarios in which safety evaluation is necessary since the performance metrics indicate a miseleading good result. The scenario is illustrated in Fig.~\ref{fig:safety_scenario}. The ego vehicle is moving towards an intersection with a reliable lane detection. However, one vehicle coming from the left which leads to a collision is not detected in this case. With ongoing frames the distance between the objects decreases, which reduces $S_{obj}$ from 0.74 in frame 1 to 0.04 in frame 28. The F1-score is not able to represent this safety critical scenario and achieves results between 0.76 and 0.93 showing a good perception. This demonstrates the general functionality of the proposed object safety metric and the necessity to apply appropriate metrics when evaluating environmental perception. There can be constructed arbitrary many scenarios in which the safety metric provide a more meaningful result compared to the performance metrics.

\subsection{Results on DeepAccident}
The results of the evaluation on the DeepAccident dataset are shown in Table~\ref{tab:results}. Considering the object detection performance metrics of precision and recall, as well as the resulting F1-score, it can be seen that most objects are correctly detected. In only a few scenarios does the recall score drop to 0.667. However, the mean for all metrics is over 0.95. The more advanced metrics, MODA and MODP, also show lower results; however, with means of 0.457 and 0.766 respectively, they are significantly lower. This can be traced back to the fact that, for these metrics, the IoU is taken as input, and the IoU is close to 0.7, leading to a high number of true positives (TPs) for precision and recall, but a lower result for the CLEAR metrics.

The contrast between safety and performance results can be seen in the CSM by Volk et al.~\cite{volk2020safety} showed a mean score of 0.510, representing severe accidents. With a minimum score of 0.00, some potential fatalities are also possible. 

Similar results to those observed for object detection can be seen for lane detection. The mean precision, recall and F1-score is between 0.82 and 0.90, with a low standard deviation. In various scenarios, the detection performance achieves perfect scores of greater than 0.99.

\begin{table}[t]
      \caption{Results on the DeepAccident dataset}
      \centering
    \begin{tabular}{l    rrrr} \toprule
    Metric& $\mu$ & $\sigma$ & min & max \\ \midrule
    \multicolumn{5}{c}{\textbf{Object Detection}}\\
    $P$ & 0.996 & 0.002& 0.990& 1.000\\
    $R$ & 0.950 & 0.024& 0.667& 1.000\\
    $F1$ & 0.974 & 0.012& 0.800& 1.000\\
    $MODA$ & 0.457 & 0.082& 0.000& 0.917\\
    $MODP$ & 0.776 & 0.064& 0.000& 0.949\\
    $CSM$ & 0.510 & 0.194& 0.000& 0.900\\
    \midrule
    \multicolumn{5}{c}{\textbf{Lane Detection}}\\
    $P$ & 0.823 & 0.122& 0.642& 0.997\\
    $R$ & 0.901 & 0.057& 0.877& 1.000\\
    $F1$ & 0.869 & 0.087& 0.800& 0.998\\
    
    \midrule
    \midrule
    \multicolumn{5}{c}{\textbf{EPSM (ours)}}\\
    $S_{obj}$ & 0.697 & 0.241& 0.019& 1.000\\
    $S_{lane}$ & 0.542 & 0.175& 0.000& 0.998\\
    $S_F$ & 0.552 & 0.163& 0.012& 0.985\\
    \bottomrule
\end{tabular}
\vspace*{0mm}
\label{tab:results}
\end{table}

For our EPSM, a similar classification as for the CSM can be observed.
Our novel object safety metric achieved a mean $S_{obj}$ of about 0.7, ranging between 0.02 and 1.00. This shows that, in most cases, the object detection is safe; however, in some cases highly critical objects are missed. The same behavior can be observed for the lane safety evaluation with $S_{lane}$ ranging between 0.0 and about 1.0, with a mean of 0.542 which represents more safety-critical situations.
The results of our final safety score $S_F$ show a mean of 0.552 with $\sigma = 0.163$. While for some frames a minimum with 0.012 representing potential fatalities exists, other frames show a perfectly safe perception achieving a safety score of 0.985. These high scores are matching some of the performance results, because a perfect detection for lanes and all objects can be considered as safe. However, the minimum indicates that highly safety-critical situations can occur, which are not covered by the performance metrics.

In general, the results of safety evaluations are similar to those of performance metrics in most scenarios, as good and bad perception correlate with high and low safety, respectively. However, there are various cases in which performance metrics showed good scores, but safety evaluation methods indicated that the perception could not be considered safe. For our EPSM, the indications based on the results are similar to those of the CSM and LSM. However, our metric considers task interdependence, which can affect safety in some cases, and is therefore more meaningful than a single evaluation using the CSM and LSM. The evaluation also demonstrates that our EPSM allows for easier evaluation, as there is a single value as result.

\section{CONCLUSION \& OUTLOOK}
\label{sec:conclusion}

In this paper we proposed a novel and highly variable metric for a joint safety evaluation of object and lane detection systems in autonomous driving. We highlighted that state-of-the-art performance metrics are not suitable to evaluate detection algorithms and can indicate misleading results in different scenarios. Unlike these performance metrics our offline safety metric allows for a more comprehensive and meaningful evaluation in terms of safety.

For this, we propose a novel lightweight object safety metric which evaluates the criticality of an object by using multi-metric aggregation and bidirectional rating strategies. In addition, the severity is taken into account by applying object type-based logistic regression models determining the effect of a potential collision including for distinction between vehicle-vehicle and vehicle-VRU collisions. This improvement in criticality classification in combination with real-world severity models allows for a more meaningful statement about the safety of object detection as other state-of-the-art metrics. This novel object safety metric is combined with an adapted version of a lane safety metric by Gamerdinger et al.~\cite{gamerdinger2024lsm} using a power mean function for a higher influence of the worse intermediate metric result. Additionally, our metric incorporates the interdependence between both tasks, as the lane detection can affect the criticality of objects.

A key advantage is that the proposed perception safety metric results in a single score $S_F\in[0,1]$ which allows an easy comparability between different methods. Moreover, the categorization into five safety stages (insufficient, very bad, bad, good and very good) allows a fast safety assessment and simplifies the interpretation of the result. Moreover, the metric can be used for single-frame detection or multi-frame detection with an object tracking and lane management system.

It is important to note that this metric cannot guarantee the safety of an autonomous vehicle, as this metric only covers the offline safety evaluation of lanes and object but considerably extends existing safety metrics and performance evaluation techniques. Thus, the metric provides an important component towards a comprehensive and composable safety evaluation for environmental perception and safe autonomous driving.

For future research, we aim to extend the metric to incorporate traffic sign recognition. Additionally, we will benchmark different combinations of lane and object detection methods in order to provided meaningful statements about their safety under diverse environmental conditions. The safety evaluation framework will be released in order to increase safety in autonomous driving.







\bibliographystyle{IEEEtran} 
\bibliography{literature.bib}

\end{document}